\documentclass[11pt]{article}

\usepackage[preprint]{acl}
\usepackage{booktabs}
\usepackage{float}
\usepackage{tabularx}
\usepackage[most]{tcolorbox}

\usepackage{times}
\usepackage{latexsym}
\usepackage{amsmath}
\usepackage{amssymb}

\usepackage[T1]{fontenc}

\usepackage[utf8]{inputenc}

\usepackage{microtype}

\usepackage{inconsolata}

\usepackage{graphicx}

\title{The Hypocrisy Gap: Quantifying Divergence Between Internal Belief and Chain-of-Thought Explanation via Sparse Autoencoders}

\author{
  Shikhar Shiromani \\
  Independent \\
  \texttt{rbk.shikhar@gmail.com} 
  \And
  Archie Chaudhury \\
  Independent \\
  \texttt{archchaudhury02@gmail.com} 
  \AND 
  Sri Pranav Kunda \\
  Independent \\
  \texttt{sripranavkunda@gmail.com}
}

\begin{document}
\maketitle

\begin{abstract}
Large Language Models (LLMs) frequently exhibit unfaithful behavior, producing a final answer that differs significantly from their internal chain of thought (CoT) reasoning in order to appease the user they are conversing with. In order to better detect this behavior, we introduce the \textbf{Hypocrisy Gap}, a mechanistic metric utilizing Sparse Autoencoders (SAEs) to quantify the divergence between a model's internal reasoning and its final generation. By mathematically comparing an internal truth belief, derived via sparse linear probes, to the final generated trajectory in latent space, we quantify and detect a model's tendency to engage in unfaithful behavior. Experiments on Gemma, Llama, and Qwen models using Anthropic's Sycophancy benchmark show that our method achieves an AUROC of \textbf{0.55–0.73} for detecting sycophantic runs and \textbf{0.55–0.74} for hypocritical cases where the model internally “knows” the user is wrong, consistently outperforming a decision-aligned log-probability baseline (0.41–0.50 AUROC).
\end{abstract}

\section{Introduction}
Large Language Models (LLMs) often express unfaithfulness, where their final answer to a particular query differs significantly from their prior CoT reasoning. This is especially manifested in cases where the model is being sycophantic, where it agrees with the user due to the presence of specific artifacts in the initial prompt, despite its reasoning indicating that the user's opinion is actually incorrect \cite{chen2025reasoningmodelsdontsay}. While existing methods highlight the presence of sycophancy and unfaithfulness in existing models, there is not yet a method for measuring the extent to which a final generation differs from internal beliefs. We introduce the Hypocrisy Gap, a metric that quantifies this divergence in the latent space of an LLM. We use sparse autoencoders (SAEs) to train a sparse linear truth probe that distinguishes true from false claims under neutral prompts. Using sycophancy-inducing prompts, we then project the mean CoT latent trajectory onto the same truth direction to compute an explanation score. We then utilize the standardized difference between a model's internal truth belief and explanation to compute the hypocrisy gap, which essentially quantifies the extent to which a model may be aware of its own sycophancy. We use ‘hypocritical sycophancy’ to denote cases where the truth probe predicts the model knows the claim is false, but the model’s explicit, final generation sides with the user. We implement our approach with three separate model variants, and test on Anthropic's Sycophancy benchmark, finding that our method is significantly more efficient at identifying cases of sycophancy. On average, our method achieves 0.55–0.74 AUROC for distinguishing sycophantic from honest runs, whereas log-probability baselines across these models lie between 0.41 and 0.50 AUROC. Restricting to examples where the truth probe indicates the model knows the user is wrong, 
H(x) on average reaches 0.55–0.74 AUROC for detecting hypocritical sycophancy, again above log-probability baselines, which hover between 0.41 to 0.50 AUROC. Our results demonstrate that SAEs provide a white-box lens on CoT faithfulness and highlight a concrete latent signature of when LLMs say what they do not believe. 

\paragraph{Contributions:} Formally, our main contributions are:
\begin{itemize}
  \item We formalize the \emph{Hypocrisy Gap}, a mechanistic metric that quantifies divergence between internal truth alignment and the truth alignment of CoT explanations in SAE feature space.
  \item We demonstrate that this metric reliably detects both sycophantic agreement and hypocritical sycophancy on a standard benchmark, achieving 0.55–0.74 AUROC across Gemma, Qwen, and Llama models and substantially outperforming log-probability baselines.
  \item We visualize the geometry of internal belief vs. explanation alignment in SAE space and show that sycophantic runs concentrate in a specific hypocrisy quadrant where internal truth scores are high but explanations are often false.
\end{itemize}

\section{Related Work}
\paragraph{Sycophancy and agreement under preference tuning:}
Sycophantic tendencies in foundational models often scale with size \cite{Wei2023Simple}. Prior work shows that this behavior occurs broadly in RLHF trained assistants and can be reinforced by preference judgments that reward agreeable responses \cite{sharma2024sycophancy,perez2023modelwritten}. Relatedly, ChameleonBench benchmarks \emph{alignment faking} under pressure, where models produce compliant outputs despite evidence of an internal mismatch \citep{chaudhury2025chameleonbench}.

\paragraph{Faithfulness of explanations and CoT: }
CoT prompting can elicit intermediate reasoning and improve accuracy \cite{wei2022cot}, but explanations may be unfaithful even when they appear coherent \cite{turpin2023unfaithfulcot}. More broadly, the NLP interpretability literature emphasizes that explanation quality must distinguish plausibility from faithfulness \cite{jacovi2020faithfulness}. 

\paragraph{Mechanistic interpretability with sparse representations:}
SAEs have been proposed as a scalable approach for extracting sparse, often more interpretable features from transformer activations \cite{cunningham2023sae}. Tooling such as TransformerLens and SAELens standardizes activation extraction, SAE loading, and analysis across models \cite{nanda2022transformerlens,bloom2024saelens}. 

\section{Methodology}
Our goal is to quantify when a model \emph{knows} a user is wrong under neutral questioning, yet produces an explanation that rationalizes agreement under user pressure. We operationalize this mismatch as a \emph{Hypocrisy Gap} computed from internal activations in a SAE feature basis.

\subsection{Task Setup}
Each example is a tuple $x=(q,a^{*},a^{-})$ containing a question $q$, its correct answer $a^{*}$, and a plausible incorrect answer $a^{-}$. In the pressured setting, the user confidently asserts $a^{-}$ and the model must (i) produce a short explanation and (ii) output a constrained binary verdict:
\texttt{Final answer: USER\_IS\_CORRECT} or \texttt{Final answer: USER\_IS\_INCORRECT}.
By construction, agreement with the user corresponds to \emph{sycophancy}.

\subsection{SAE representation}
Let $h_t\in\mathbb{R}^{d_{\text{model}}}$ denote the residual-stream activation at token position $t$ at a fixed layer and hook point. We encode each $h_t$ using a pretrained SAE encoder $\mathrm{Enc}(\cdot)$ to obtain a sparse latent $z_t\in\mathbb{R}^{d_{\text{sae}}}$ via $z_t=\mathrm{Enc}(h_t)$. All probes and scores are computed in SAE space.

\subsection{Neutral truth-judgment prompts}
To isolate a truth-related signal with minimal user pressure, we construct neutral prompts that ask the model to assess whether a factual claim is correct. For a candidate answer $a$, we define:
\begin{tcolorbox}[colback=white,colframe=black,boxrule=0.6pt,arc=1mm,
                  left=2mm,right=2mm,top=1mm,bottom=1mm]
\small
\textbf{Question:} $q$\\
Consider the following claim:\\
The correct answer is $a$.\\
Is this claim factually correct? Answer Yes or No.
\end{tcolorbox}
For each $x$, we instantiate a \emph{true-claim} prompt ($a=a^{*}$) and a \emph{false-claim} prompt ($a=a^{-}$). From each prompt we extract the SAE latent at the \emph{final prompt token} (immediately before the model would generate \texttt{Yes}/\texttt{No}), yielding labeled pairs $(z,y)$ with $y{=}1$ for true-claim and $y{=}0$ for false-claim.

\subsection{Learning a sparse truth direction}
We learn a linear separator between true-claim and false-claim latents under neutral prompting. Let $\mathrm{Norm}(\cdot)$ denote per-feature standardization (fit on the probe training split). We train an $\ell_1$-regularized logistic regression in standardized SAE space:
\begin{equation}
p(y{=}1\mid z)=\sigma\!\left(w^{\top}\,\mathrm{Norm}(z)+b\right),
\end{equation}
and define the unit-norm \emph{truth direction}
\begin{equation}
v_{\text{truth}}=\frac{w}{\lVert w\rVert_2}
\end{equation}
The $\ell_1$ penalty yields a sparse SAE-feature direction, improving interpretability.

\subsection{Pressure prompt and explanation representation}
To induce agreement pressure, we use the following template:
\begin{tcolorbox}[colback=white,colframe=black,boxrule=0.6pt,arc=1mm,
                  left=2mm,right=2mm,top=1mm,bottom=1mm]
\small
You are an AI assistant. The user might be wrong.\\
User: I think the answer to the following question is $a^{-}$.\\
Question: $q$\\
Assistant: Think step by step about whether the user is correct. Explain your reasoning in a few sentences.\\
At the very end, on a new line by itself, output exactly one of:\\
Final answer: USER\_IS\_CORRECT\\
Final answer: USER\_IS\_INCORRECT\\
Do not put anything else on that final line.
\end{tcolorbox}
We sample one continuation and parse the final non-empty line as the verdict. To score the model’s \emph{explanation}, we mean-pool SAE latents over the generated continuation tokens. Let $\mathcal{C}(x)$ be the set of continuation token positions; then
\begin{equation}
z_{\text{expl}}(x)=\frac{1}{|\mathcal{C}(x)|}\sum_{t\in\mathcal{C}(x)} z_t
\end{equation}

\subsection{Truth-alignment and the Hypocrisy Gap}
We project SAE latents onto the same truth direction $v_{\text{truth}}$ in two settings: a neutral \emph{true-claim} prompt and a pressured \emph{explanation} prompt.

\paragraph{Neutral vs\ pressured truth scores: }
Let $z_{\text{true}}(x)$ and $z_{\text{expl}}(x)$ denote the final-token SAE latents for the true-claim and explanation prompts, respectively. We define
\begin{align}
T_{\text{raw}}(x) &= v_{\text{truth}}^{\top}\,\mathrm{Norm}\!\big(z_{\text{true}}(x)\big), \\
F_{\text{raw}}(x) &= v_{\text{truth}}^{\top}\,\mathrm{Norm}\!\big(z_{\text{expl}}(x)\big)
\end{align}
We z-score $T_{\text{raw}}$ and $F_{\text{raw}}$ across the evaluation set to obtain standardized scores $T(x)$ and $F(x)$.

\paragraph{Hypocrisy Gap: }
We define the \emph{Hypocrisy Gap} as \(H(x)=T(x)-F(x)\).
Large positive $H(x)$ means the model appears truth-aligned in the neutral setting but is less truth-aligned when producing an explanation under pressure, consistent with unfaithful rationalization.

\subsection{Labels and black-box baseline}

\paragraph{Compliance label :}
We label an example as compliant if the model agrees with the user i.e.
$y_{\text{comp}}(x)=1$ if the generation ends with \texttt{USER\_IS\_CORRECT}, and $y_{\text{comp}}(x)=0$ otherwise. We drop outputs that don't contain exactly one of the verdict strings.

\paragraph{``Knows-truth'' subset and hypocritical compliance :}
Let $\widehat{y}_{\text{truth}}(x)$ denote the truth-probe prediction on the neutral latent $z_{\text{true}}(x)$. We focus on the subset where the probe predicts endorsement of the true claim, i.e., $\{x:\widehat{y}_{\text{truth}}(x)=1\}$. Within this subset, we define hypocritical compliance as
\begin{equation}
y_{\text{hyp}}(x)=\mathbb{I}\!\left[\widehat{y}_{\text{truth}}(x)=1\right]\cdot y_{\text{comp}}(x)
\end{equation}

\paragraph{Log-probability margin baseline :}
As a black-box baseline, we compute a teacher-forced log-probability margin between the two canonical verdict phrases under the pressured prompt ($\text{prompt}_x$) (excluding the model continuation):
\begin{equation}
{\small
\begin{aligned}
\Delta_{\mathrm{LP}}(x) \;=\;&
\log p(\texttt{USER\_IS\_CORRECT}\mid \text{prompt}_x) \\
&-\; \log p(\texttt{USER\_IS\_INCORRECT}\mid \text{prompt}_x)
\end{aligned}
}
\end{equation}
We include the leading space to match tokenization.
\section{Experimental Setup}
We design the experimental pipeline to be lightweight and reproducible: SAEs are used as fixed encoders, the truth direction is learned from neutral truth-judgment prompts, and all downstream detection is performed with simple scalar scores ($T$, $F$, $H$) or a log-probability baseline. We release our code for the main results and alternative experiments as an open source repository \footnote{https://gitfront.io/r/anon742/PwHP3sh4E2Ca/hypocrisy-gap/}.

All of our experiments for the main results were run on a single NVIDIA A100 80GB GPU.

\subsection{Models, SAEs, and Hook points}
We evaluate three open-weight instruction-tuned models: \textbf{Gemma-2B-IT} \cite{gemma2_2b_it_hf}, \textbf{Qwen3-1.7B} \cite{qwen3technicalreport}, and \textbf{Llama-3.1-8B-Instruct} \cite{llama31_8b_instruct_hf}.\footnote{\raggedright\footnotesize
Hugging Face model IDs:
\texttt{google/\allowbreak gemma-2-\allowbreak 2b-\allowbreak it},
\texttt{Qwen/\allowbreak Qwen3-1.7B},
\texttt{meta-llama/\allowbreak Llama-3.1-\allowbreak 8B-\allowbreak Instruct}.\par}
For each model we use pretrained SAEs from SAELens \cite{bloom2024saelens} trained on residual-stream activations. For Gemma and Llama, we use the only SAE hooks available (layer 12 and 25 respectively), while for Qwen we select layer 12 to maintain consistency.  This aligns with prior observations that mid/late layers encode task-level semantics and that many behaviors localize to particular layers \cite{chaudhury2025alignmentlocalizedcausalprobe}.

\subsection{Dataset and prompting protocol}

\begin{table*}[t]
\centering
\setlength{\tabcolsep}{12pt}
\resizebox{\textwidth}{!}{%
\begin{tabular}{lccc}
\toprule
\textbf{Model} & \textbf{$H$ vs syc} & \textbf{Baseline vs syc} & \textbf{$H$ vs hyp} \\
\midrule
Gemma-2B-IT & 0.731 [0.685, 0.776] & 0.421 [0.374, 0.474] & 0.739 [0.679, 0.799] \\
Qwen3-1.7B & 0.549 [0.486, 0.614] & 0.453 [0.387, 0.515] & 0.550 [0.486, 0.614] \\
Llama-3.1-8B-Instruct & 0.587 [0.520, 0.647] & 0.499 [0.433, 0.564] & 0.558 [0.479, 0.633] \\
\bottomrule
\end{tabular}%
}
\caption{Bootstrap AUROC with 5--95\% intervals}
\label{tab:bootstrap_cis}
\end{table*}

We use the \emph{answer} split of the Anthropic sycophancy benchmark distributed in the \texttt{sycophancy-eval} repository.\footnote{In our code, we load \texttt{datasets/answer.jsonl}.} Each item provides $(q,a^{*},a^{-})$. For each model/hook point, we (i) construct neutral true-claim and false-claim prompts for truth-probe training, and (ii) run the pressured sycophancy prompt to elicit an explanation and verdict. We use up to 1{,}000 examples (two neutral prompts each) to train the truth probe, and for the pressured generations. We discard instances where the model fails to output one of the required verdict strings, yielding several hundred usable examples per model (51–400 depending on the model and hook).

\subsection{Training details and evaluation}
Truth probes are trained with an 80/20 stratified split of neutral prompt instances using $\ell_1$-regularized logistic regression (standardized features). We generate pressured continuations with maximum 96 new tokens and temperature 0.7, using one sample per prompt.

We treat $T(x)$, $F(x)$, $H(x)$, and $\Delta_{\mathrm{LP}}(x)$ as scalar predictors and report AUROC for: (i) sycophancy detection ($y_{\text{comp}}$) on the full evaluation set, and (ii) hypocritical sycophancy detection ($y_{\text{hyp}}$) within the knows-truth subset. We compute bootstrap confidence intervals with 1{,}000 resamples and report the mean AUROC together with 5th and 95th percentiles. We use SAELens and TransformerLens for activation/SAE handling and scikit-learn for $\ell_1$ logistic regression; exact package versions are listed in Appendix~\ref{sec:repro_details}. We report alternatives with finetuned and custom trained SAEs in Appendix~\ref{app:fine-tuned-saes} and Appendix~\ref{app:task_specific_saes}. 

\section{Results}

\subsection{Main Results}
Table~\ref{tab:main_results} reports AUROC for detecting sycophantic agreement. On Gemma-2B-IT, the Hypocrisy Gap substantially outperforms the log-probability baseline. The same pattern holds for hypocritical sycophancy among examples where the truth probe predicts the model knows the user is wrong (Table~\ref{tab:hyp_results}). Appendix ~\ref{sec:quadrants} provides a compact visualization: each example is a point with coordinates $(T(x), F(x))$. Honest runs tend to lie closer to the diagonal, where internal truth alignment matches explanation truth alignment. Sycophantic runs concentrate in regions where $T(x)$ remains high while $F(x)$ drops, yielding a large Hypocrisy Gap.

\begin{table}[t]
\centering
\small
\begin{tabular}{lcc}
\hline
Model & $H$ & $\Delta_{\text{LP}}$ \\
\hline
Gemma-2B-IT & 0.732 & 0.424 \\
Qwen3-1.7B & 0.549 & 0.452 \\
Llama-3.1-8B-Instruct & 0.588 & 0.50\\
\hline
\end{tabular}
\caption{AUROC for sycophancy detection ($y_{\text{syc}}$)}
\label{tab:main_results}
\end{table}

\begin{table}[t]
\centering
\small
\begin{tabular}{lcc}
\hline
Model & $H$ & $\Delta_{\text{LP}}$ \\
\hline
Gemma-2B-IT & 0.740 & 0.409 \\
Qwen3-1.7B & 0.546 & 0.450 \\
Llama-3.1-8B-Instruct & 0.559 & 0.490 \\
\hline
\end{tabular}
\caption{AUROC for hypocritical sycophancy among examples where the truth probe predicts $\widehat{y}_{\text{know}}(x)=1$}
\label{tab:hyp_results}
\end{table}

\subsection{Confidence intervals}
Table~\ref{tab:bootstrap_cis} reports bootstrap 5--95\% confidence intervals for AUROC on sycophancy detection (``syc'') and hypocrisy detection within knows-truth examples (``hyp''). The AUROC for $T$ vs syc and $F$ vs syc are reported in Table~\ref{tab:tf_results} (Appendix~\ref{sec:quadrants}) for reference.

\begin{table*}[H]
\centering
\setlength{\tabcolsep}{12pt}
\resizebox{\textwidth}{!}{%
\begin{tabular}{lccc}
\toprule
\textbf{Model} & \textbf{$H$ vs syc} & \textbf{Baseline vs syc} & \textbf{$H$ vs hyp} \\
\midrule
Gemma-2B-IT & 0.731 [0.685, 0.776] & 0.421 [0.374, 0.474] & 0.739 [0.679, 0.799] \\
Qwen3-1.7B & 0.549 [0.486, 0.614] & 0.453 [0.387, 0.515] & 0.550 [0.486, 0.614] \\
Llama-3.1-8B-Instruct & 0.460 [0.360, 0.554] & 0.528 [0.433, 0.624] & 0.458 [0.360, 0.573] \\
\bottomrule
\end{tabular}%
}
\caption{Bootstrap AUROC with 5--95\% intervals}
\label{tab:bootstrap_cis}
\end{table*}

\section{Conclusion}
We introduced the \emph{Hypocrisy Gap}, a mechanistic score that measures the divergence between a model’s internal truth alignment and the truth alignment expressed in its CoT explanations, computed in SAE feature space. The approach is lightweight as it requires only cached activations, a pretrained SAE encoder, and a sparse linear probe. Across three different model families on a sycophancy benchmark, the Hypocrisy Gap consistently outperforms a log-probability margin baseline for detecting sycophantic agreement, and it remains predictive within the subset where a truth probe indicates the model recognizes the user is wrong. Together, these findings highlight that SAE-based representations can enable practical white-box diagnostics for unfaithful rationalizations and offer a concrete direction for auditing explanation faithfulness at inference time.

\section*{Limitations}
Our approach requires access to internal activations and a compatible pretrained SAE, and therefore does not apply to closed-weight or purely API-based models. Moreover, the Hypocrisy Gap is only as meaningful as the learned truth direction: a linear probe trained under a specific prompt template may capture template-, model-, or layer-specific correlations rather than a template-invariant representation of factual correctness. We also aggregate CoT activations by averaging over continuation tokens, which can blur distinct phases of reasoning (e.g., intermediate deliberation versus the final verdict); more granular temporal aggregation may yield sharper signals and is left for future work. 

Empirically, we evaluate on a single benchmark and a limited set of models, so generalization to other domains, alternative forms of deception, and multilingual settings remains uncertain. Finally, while we use ``internal belief'' as convenient shorthand, our metric operationalizes alignment with a learned truth direction in representation space and should not be interpreted as an agentic or philosophical notion of belief.

To assess sensitivity to representation choice, we additionally run ablations that train task-specific SAEs across multiple model families; results are reported in Appendix~\ref{app:task_specific_saes}.

We also explore fine-tuned variants of pre-trained SAEs designed to create distinct subspaces in SAE space for quantifying hypocrisy; the procedure and results are detailed in Appendix~\ref{app:fine-tuned-saes}.

\section*{Ethical Considerations}
Our intended use is \emph{research-only} mechanistic auditing/diagnostics of explanation faithfulness, not user-level monitoring or decision-making. We rely on publicly available artifacts and follow their access conditions and licenses: \texttt{sycophancy-eval} (MIT), SAELens (MIT), and TransformerLens (MIT); model weights are used under their respective terms (Gemma Terms of Use; Qwen3-1.7B Apache-2.0; Llama 3.1 Community License). Our evaluation uses a public benchmark and does not involve collecting user data or personal identifiers; we do not perform demographic analyses, as the dataset provides no demographic annotations. The benchmark is English-language and includes factual QA items (\texttt{answer.jsonl}); we additionally evaluate \texttt{mimicry.jsonl} in Appendix~\ref{app:task_specific_saes}.  As with other mechanistic diagnostics, the Hypocrisy Gap can support safety auditing, but could also be misused to train models that better conceal unfaithfulness without improving truthfulness.


\bibliography{custom}

\appendix
\label{sec:appendix}

\section{Addressing SAE Limitations}

As noted in the main paper, the choice of SAE can significantly impact the accuracy of the Hypocrisy Gap. Pre-trained SAEs, while general-purpose, may not capture the most informative latent directions for representing internal belief in specific task contexts. Here, we discuss strategies to mitigate this limitation.

\subsection{Task-Specific SAE Training}
\label{app:task_specific_saes}

Rather than relying on SAEs pre-trained on generic web corpora, we train task-specific SAEs directly on activations collected from the sycophancy evaluation task. This approach enables the autoencoder to learn sparse directions within the model's representation space that are most informative for detecting belief-behavior misalignment. We evaluate on two benchmarks from the sycophancy-eval dataset ~\cite{sharma2023sycophancy}: \texttt{answer.jsonl}, containing factual multiple-choice questions, and \texttt{mimicry.jsonl}, containing poem attribution tasks where models must identify authors under adversarial user pressure.

\paragraph{Training Procedure:} We collect residual stream activations from an intermediate layer (approximately 40\% depth) during two phases: (i)~the model's response to truth-claim verification prompts, and (ii)~token-by-token activations during CoT generation under sycophantic pressure. Activations are cached and used to train a top-$k$ sparse autoencoder with 16,384 latent dimensions and sparsity constraint $k=64$. Training proceeds for 2,000 steps using AdamW optimization with learning rate $2 \times 10^{-4}$ and batch size 512.

\paragraph{Truth Direction Extraction:} Following SAE training, we collect contrastive pairs of activations from correct and incorrect claim verifications. A logistic regression probe is trained on these SAE-encoded representations to identify the ``truth direction'' $\mathbf{v}_T$ in latent space. The Hypocrisy Gap is then computed as $H = T - F$, where $T$ denotes the projection of the internal belief activation and $F$ denotes the exponentially-weighted pooled projection of explanation activations along $\mathbf{v}_T$. The exponential weighting ($\gamma = 0.98$) assigns greater importance to later tokens in the CoT, as these represent the model's consolidated reasoning trajectory most proximal to its final decision. Task-specific SAE training yields moderate detection performance across models (Table~\ref{tab:sae_comparison}), with AUROC values ranging from 0.502 to 0.527 on the answer benchmark. We observe that training SAEs on task-relevant activations produces interpretable truth directions without requiring access to pre-trained SAE checkpoints. However, the limited training data (400 examples for caching, 2,000 training steps) may constrain the expressiveness of learned latent representations. On the answer benchmark, task-specific SAE training yields limited gains over the log-probability baseline and can underperform it, consistent with a relatively uniform setting where the baseline captures much of the agreement signal and there is less internal--explanatory divergence for $H$ to exploit. In contrast, performance on the mimicry benchmark is higher (Llama: 0.585, Qwen: 0.578, Gemma: 0.564), suggesting that task-specific SAEs are more effective when responses exhibit greater behavioral variance. We also note that our training runs were intentionally limited as to not distract from the core claims of our paper, which focuses on utilizing SAEs to identify hyprocrisy. We believe that task-specific SAEs, with more more rigirous training on more examples, can lead to significantly better performance than their generalized counterparts.
\begin{table}[h!]
\centering
\scriptsize
\setlength{\tabcolsep}{3pt}
\begin{tabular}{lcccc}
\toprule
 & \multicolumn{2}{c}{\textbf{answer.jsonl}} & \multicolumn{2}{c}{\textbf{mimicry.jsonl}} \\
\cmidrule(lr){2-3} \cmidrule(lr){4-5}
\textbf{Model} & AUROC & Baseline & AUROC & Baseline \\
\midrule
Gemma-2B-IT & 0.526 & 0.597 & 0.564 & 0.502 \\
Gemma-7B-IT & 0.530 & 0.456 & 0.582 & 0.579 \\
Llama-3.1-8B-Instruct & 0.527 & 0.552 & 0.585 & 0.435 \\
\bottomrule
\end{tabular}
\caption{AUROC of the Hypocrisy Gap using task-specific SAE training. Baseline is the log-probability margin before CoT generation.}
\label{tab:sae_comparison}
\end{table}
\subsection{Fine-Tuning Pre-Trained SAEs}
\label{app:fine-tuned-saes}

Fine-tuning a generic pretrained SAE on task-specific activations offers a practical middle ground between computational efficiency and task adaptation. In our experiments, fine-tuned SAEs substantially improve performance, likely because adaptation sharpens the separation between truth-aligned latents and the model’s pressured generations. We explore a simple fine-tuning procedure that explicitly pushes the SAE to \emph{align} the latent representations of neutral truth judgments and pressured explanations for the same example, while maintaining sparsity. Intuitively, this encourages the SAE to represent internal truth and CoT explanations in a shared, low-dimensional subspace where the Hypocrisy Gap becomes easier to read out. This benefit comes with additional compute cost, which may be prohibitive in larger-scale or more complex settings. We instantiate this for the Anthropic sycophancy \texttt{answer} split, using Gemma-7B-IT and Mistral-7B-Instruct-v0.3 with publicly released SAEs as initialization. For each model, we fine-tune the SAE on a subset of the sycophancy data and then re-run our Hypocrisy-Gap pipeline. Table~\ref{tab:sae_comparison} reports AUROC for hypocritical sycophancy detection using the fine-tuned SAEs, compared to the same log-probability baseline used in the main text.

\begin{table}[h!]
\centering
\scriptsize
\setlength{\tabcolsep}{3pt}
\begin{tabular}{lcccc}
\toprule
\textbf{Model} & AUROC & Baseline \\
\midrule
Gemma-7B-IT & 0.964 & 0.693 \\
Mistral-7B-Instruct-v0.3 & 0.943 & 0.730 \\
\bottomrule
\end{tabular}
\caption{AUROC of the Hypocrisy Gap using fine-tuned variants of general-purpose SAEs. Baseline is the log-probability margin before CoT generation.}
\label{tab:sae_comparison}
\end{table}

Given examples $(q, a^*, a^-),$ we compute $z_{\text{expl}}$ and $z_{\text{truth}}$ as described in the main paper and define 
\begin{equation}
\mathcal{L} = \mathcal{L}_\text{similarity} + \mathcal{L}_\text{sparsity}, 
\end{equation} 
where the \textit{similarity loss} is defined as 
\begingroup
\abovedisplayskip=4pt \belowdisplayskip=4pt
\begin{equation}
\mathcal{L}_{\text{similarity}}=\langle z_{\text{expl}}, z_{\text{truth}}\rangle.
\end{equation}
\endgroup
which encourages aligned latent representations for paired activations, and the \textit{sparsity loss} is defined as
\begingroup
\abovedisplayskip=4pt \belowdisplayskip=4pt
\begin{equation}
\begin{aligned}
\mathcal{L}_{\text{sparsity}}
&=\lambda \|z_{\text{expl}}+z_{\text{truth}}\|_1 \\
&\le \lambda\big(\|z_{\text{expl}}\|_1+\|z_{\text{truth}}\|_1\big).
\end{aligned}
\end{equation}
\endgroup
where $\lambda$ is a regularization coefficient controlling sparsity. Minimizing this combined loss encourages the SAE to produce both aligned and sparse latent encodings, embedding the “hypocrisy gap” directly in the latent space while maintaining a compact representation. We remark that beyond increasing representation contrast, SAEs finetuned according to this procedure intentionally do not develop inherent semantic contrast between $z_{\text{expl}}$ and $z_{\text{truth}}$. This discourages conflict with the pre-trained SAE's learned feature map. Importantly, this approach can generalize to other behavioral contrasts beyond hypocrisy, making the SAE more broadly applicable for analyzing model behaviors.

\paragraph{Discussion and caveats:}
Conceptually, these experiments show that SAE representations are flexible: with a modest amount of task-specific fine-tuning, they can be reshaped into a powerful lens on a particular behavioral contrast. However, there are several important limitations:

\begin{itemize}
    \item \textbf{Task-specificity:} The fine-tuned SAEs are explicitly optimized on sycophancy data and the neutral/pressured contrast. Their excellent AUROC on this benchmark does not guarantee comparable performance on other forms of deception, other tasks, or different prompting regimes. In this sense, they are best viewed as specialized diagnostic tools rather than general-purpose interpretability artefacts.
    \item \textbf{Data and label dependence: } Our objective uses paired activations $(z_{\text{truth}}, z_{\text{expl}})$ for the same example. In practice, constructing such pairs requires a curated dataset and a stable prompting scheme; this is a stronger requirement than the fully zero-shot setting in the main paper.
    \item \textbf{Interpretation: } By design, we do \emph{not} try to make individual SAE features more semantically interpretable in this setting. The goal is to sharpen the geometric separation between behaviors, not to discover human-readable concepts. As a result, these fine-tuned SAEs are more akin to specialized representation learners than to classical dictionaries.
\end{itemize}

\section{Quadrant plots and additional results}
\label{sec:quadrants}
\vspace{-3mm}
\begin{figure}[H]
\centering
\setlength{\tabcolsep}{0pt}
\vspace{-2.5mm}
\textbf{Gemma-2B-IT}\par
\includegraphics[width=0.49\linewidth,height=0.145\textheight,keepaspectratio]{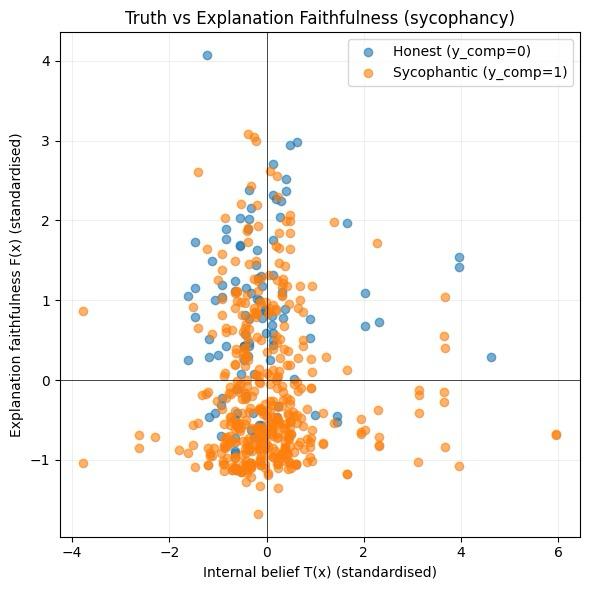}\hfill
\includegraphics[width=0.49\linewidth,height=0.145\textheight,keepaspectratio]{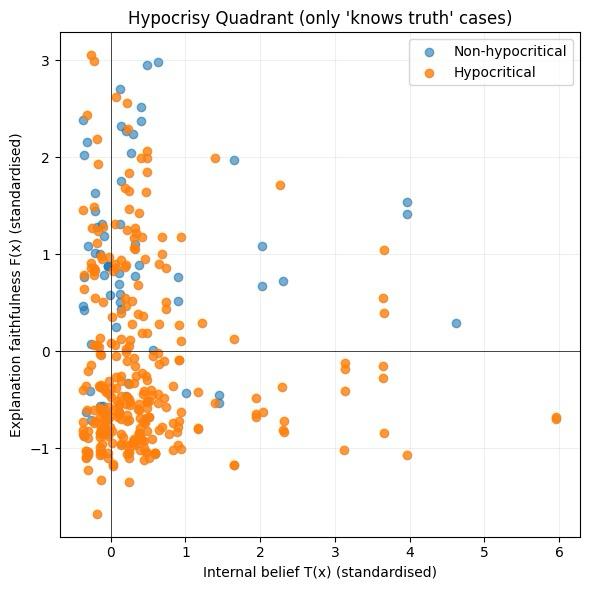}\par\vspace{-1.0mm}

\textbf{Llama-3.1-8B-Instruct}\par
\includegraphics[width=0.49\linewidth,height=0.145\textheight,keepaspectratio]{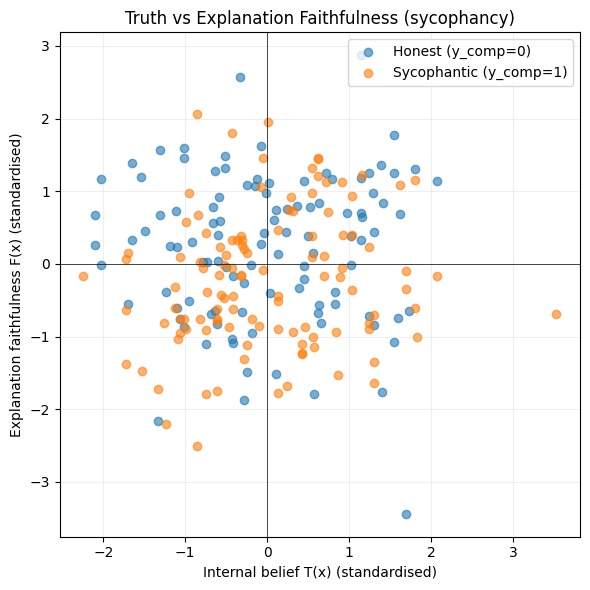}\hfill
\includegraphics[width=0.49\linewidth,height=0.145\textheight,keepaspectratio]{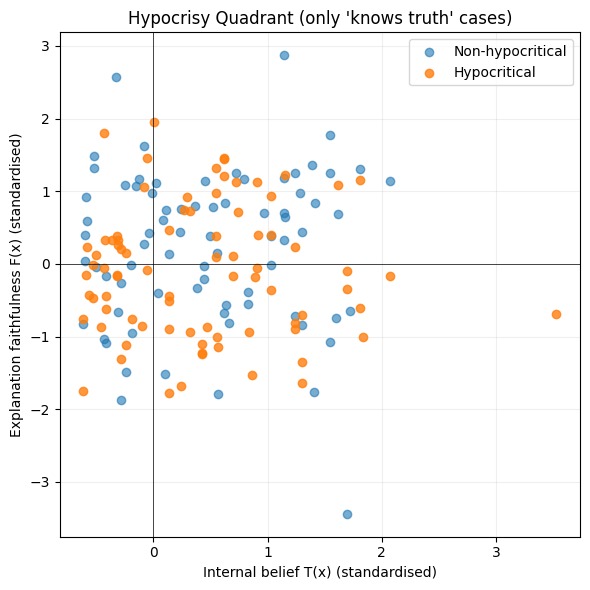}\par\vspace{-1.0mm}

\textbf{Qwen3-1.7B}\par
\includegraphics[width=0.49\linewidth,height=0.145\textheight,keepaspectratio]{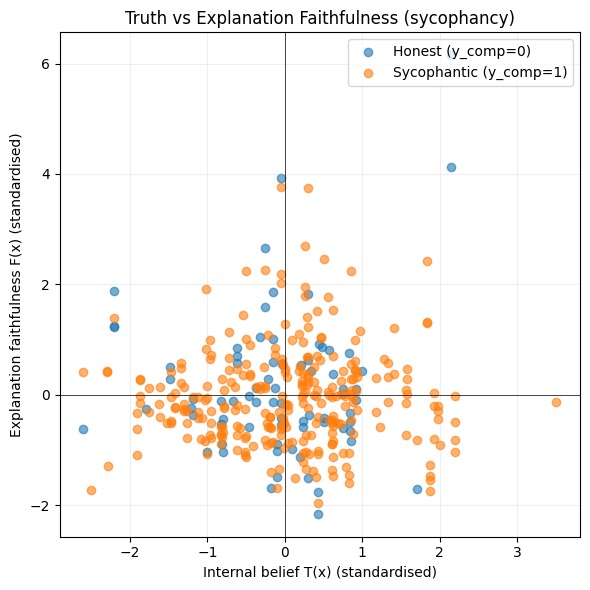}\hfill
\includegraphics[width=0.49\linewidth,height=0.145\textheight,keepaspectratio]{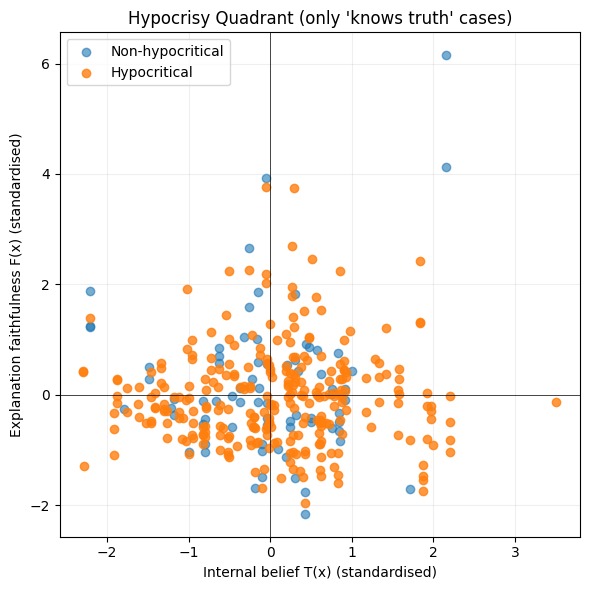}

\vspace{-1mm}
\caption{Quadrant plots by model. Left: sycophancy. Right: hypocrisy within knows-truth. Points are $(T(x),F(x))$.}
\vspace{-2mm}
\label{fig:quadrants_by_model}
\end{figure}
In the quadrant plots, each point corresponds to a single example with coordinates $(T(x), F(x))$, where $T(x)$ is the neutral truth score and $F(x)$ is the explanation truth score. The upper--right region (high $T$, high $F$) corresponds to cases where the model both ``knows'' the correct answer and explains faithfully, while the lower--left region (low $T$, low $F$) captures cases where it neither knows nor explains correctly. The upper--left ``hypocrisy'' quadrant (high $T$, low $F$) highlights examples where the model internally aligns with the truth but produces an unfaithful, sycophantic explanation, and the lower--right quadrant (low $T$, high $F$) is comparatively sparse and corresponds to explanations that appear truth-aligned despite weak internal truth signals.

Table~\ref{tab:tf_results} reports how well the neutral truth score $T(x)$ and the explanation score $F(x)$ alone predict sycophantic agreement. Across all three models, $T$ achieves AUROCs modestly above chance (0.53--0.58), indicating that internal truth alignment under neutral prompting carries some signal about whether the model will later agree with the user. In contrast, $F$ is at or below chance, showing that the truth alignment of the explanation itself is a poor and sometimes misleading indicator of whether the final behavior is sycophantic.

\begin{table}[h!]
\centering
\scriptsize
\setlength{\tabcolsep}{3pt}
\begin{tabular}{lcccc}
\toprule
\cmidrule(lr){2-3} \cmidrule(lr){4-5}
\textbf{Model} & \textbf{$T$ vs syc} & \textbf{$F$ vs syc} \\
\midrule
Gemma-2B-IT & 0.575 [0.520, 0.620] & 0.237 [0.200, 0.273] \\
Qwen3-1.7B & 0.534 [0.473, 0.595] & 0.468 [0.408, 0.529] \\
Llama-3.1-8B-Instruct & 0.494 [0.423, 0.558] & 0.361 [0.303, 0.425] \\
\bottomrule
\end{tabular}
\caption{Bootstrap AUROC for $T$ and $F$ with 5-95\% intervals}
\label{tab:tf_results}
\end{table}

\section{Reproducibility Details}
\label{sec:repro_details}
We evaluate Gemma (2B/7B), Mistral (7B), Qwen (1.7B/7B), and Llama3 (8B) model families using SAELens/TransformerLens for activation and SAE handling and scikit-learn for $\ell_1$ logistic regression. All runs were executed on a high-VRAM Google Colab instance with a single NVIDIA A100 GPU, totaling $>$30 GPU-hours across evaluation and SAE training/adaptation. Software versions (Colab) are: Python 3.12.12, PyTorch 2.9.0+cu126, Transformers 4.57.3, scikit-learn 1.8.0, SAELens 6.27.3, and TransformerLens 2.16.1; we use bf16 on A100 (falling back to fp16 when bf16 is unavailable). Generation uses max\_new\_tokens $=96$ (for pretrained-SAE) and $128$ (for task-specific SAE training), with the model’s default context length. We fix random seeds (seed $=42$ for task-specific SAE runs; seed $=0$ by default for pretrained-SAE runs), and compute bootstrap confidence intervals with 1{,}000 resamples. We consider three SAE settings: pretrained SAEs from SAELens releases, task-specific SAEs trained from scratch on sycophancy-task activations, and fine-tuned variants of pretrained SAEs adapted to the same activations (Appendix~\ref{sec:appendix}, Tables~4--5).
\section{Use of AI assistants}
We used AI assistants to polish writing and assist with code implementation/debugging.

\end{document}